\documentclass[10pt,journal,compsoc]{IEEEtran}
\renewcommand{\baselinestretch}{0.98}
\ifCLASSOPTIONcompsoc
  \usepackage[nocompress]{cite}
\else
  \usepackage{cite}
\fi
\usepackage{rotating}
\usepackage{graphicx}
\usepackage{comment}
\usepackage{url}
\usepackage{amsmath,amssymb} 
\usepackage{color}

\usepackage{multirow}
\usepackage{multicol}
\usepackage{floatrow}
\usepackage[noend]{algpseudocode}
\usepackage{algorithmicx,algorithm}
\usepackage{bbding}
\usepackage{tikz,xcolor,hyperref}
\usepackage{subfigure}
\usepackage{booktabs}
\usepackage{makecell}
\usepackage{xspace}

\definecolor{lime}{HTML}{A6CE39}
\DeclareRobustCommand{\orcidicon}{%
    \begin{tikzpicture}
    \draw[lime, fill=lime] (0,0) 
    circle [radius=0.16] 
    node[white] {{\fontfamily{qag}\selectfont \tiny ID}};    \draw[white, fill=white] (-0.0625,0.095) 
    circle [radius=0.007];    \end{tikzpicture}
    \hspace{-2mm}}
\foreach \x in {A, ..., Z}{%
    \expandafter\xdef\csname orcid\x\endcsname{\noexpand\href{https://orcid.org/\csname orcidauthor\x\endcsname}{\noexpand\orcidicon}}
    }

\begin{document}
%
\title{ \textcolor{black}{Evaluating and Mitigating Bias in AI-Based Medical Text Generation}}
%
\author{Xiuying Chen\thanks{*Equal contribution.}$^{1*\dagger}$, Tairan Wang$^{2*}$, Juexiao Zhou$^{2}$, Zirui Song$^{1}$, Xin Gao$^{2\dagger}$, Xiangliang Zhang$^{3,2\dagger}$\thanks{$\dagger$Correspondingly authors: xiuying.chen@mbzuai.ac.ae, xin.gao@kaust.edu.sa, xzhang33@nd.edu}
\thanks{
$^1$Mohamed bin Zayed University of Artificial Intelligence\\
$^2$King Abdullah University of Science and Technology\\
$^3$University of Notre Dame, IN, USA\\
}}

\markboth{}%
%
\\
\IEEEtitleabstractindextext{%
\begin{abstract}
\textcolor{black}{Artificial intelligence (AI) systems, particularly those based on deep learning models, have increasingly achieved expert-level performance in medical applications.}
However, there is growing concern that such AI systems may reflect and amplify human bias, and reduce the quality of their performance in historically under-served populations. 
The fairness issue has attracted considerable research interest in the medical imaging \textit{classification} field, yet it remains understudied in the \textit{text generation} domain. 
In this study, we investigate the fairness problem in text generation within the medical field and observe significant performance discrepancies across different races, sexes, and age groups, including intersectional groups, various model scales, and different evaluation metrics.
To mitigate this fairness issue, we propose an algorithm that selectively optimizes those underperformed groups to reduce bias. 
The selection rules take into account not only word-level accuracy but also the pathology accuracy to the target reference, while ensuring that the entire process remains fully differentiable for effective model training.
Our evaluations across multiple backbones, datasets, and modalities demonstrate that our proposed algorithm enhances fairness in text generation without compromising overall performance. 
Specifically, the disparities among various groups across different metrics were diminished by more than 30\% with our algorithm, while the relative change in text generation accuracy was typically within 2\%. 
By reducing the bias generated by deep learning models, our proposed approach can potentially alleviate concerns about the fairness and reliability of text generation diagnosis in medical domain.
 Our code is publicly available to facilitate further research at \url{https://github.com/iriscxy/GenFair}.

\end{abstract}

}

\maketitle

\IEEEdisplaynontitleabstractindextext

%
\IEEEpeerreviewmaketitle

\section{Introduction}
\textcolor{black}{Artificial Intelligence, particularly those based on deep learning models, has been widely adopted in healthcare, consistently demonstrating expert-level performance across various domains, presenting a clear incentive for real-world deployment due to the global medical expert shortage and AI algorithms matching specialist performance~\cite{jiang2017artificial,rajpurkar2017chexnet,lin2022automated,rimmer2017radiologist,chen2024scholarchemqa,wang2024nature}.}
However, the issue of fairness has arisen in medical image classification tasks, with biases observed in deep learning models related to race~\cite{obermeyer2019dissecting,seyyed2020chexclusion,zongmedfair}, sex~\cite{lin2023improving,larrazabal2020gender}, and age~\cite{lin2023improving}.
 The bias also exists in models trained from different types of medical data, such as chest X-rays \cite{seyyed2020chexclusion},
CT scans \cite{zhou2021radfusion}, and skin dermatology images \cite{kinyanjui2020fairness}.
For instance, chest X-ray classifiers trained to predict the presence of disease systematically underdiagnose black patients~\cite{lin2023evaluate}, potentially leading to delays in care.
\textcolor{black}{A biased decision-making system is socially and ethically detrimental, especially in life-changing scenarios such as healthcare~\cite{saarni2008ethical,grote2020ethics}.}

\begin{figure}[!htb]
    \centering
    \includegraphics[width=1\linewidth]{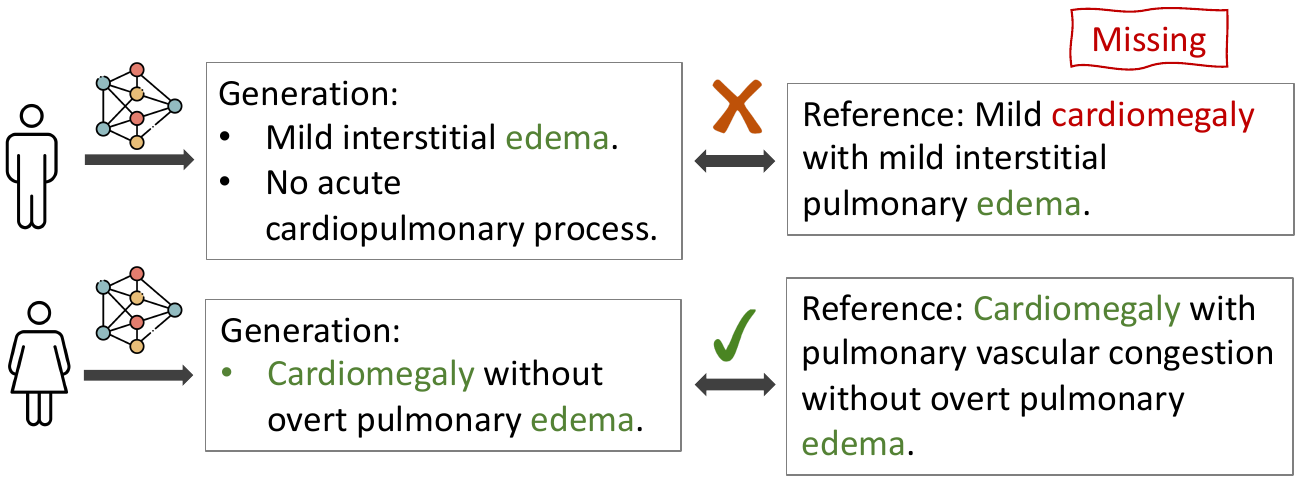}
    \caption{\textbf{An example where both male and female have similar input, yet the text generation model neglects cardiomegaly in the male case.}}
    \label{intro_case}
\end{figure}

This has motivated a growing body of work to understand bias and pursue fairness in image classification tasks~\cite{zhang2022improving,lahoti2020fairness,narasimhan2020pairwise,yang2024limits}.
For example, \cite{lin2023improving} proposed an algorithm that leverages the marginal pairwise equal opportunity to reduce bias in medical image classification. 
However, the fairness of text generation in medical contexts remains largely underexplored. 
This is particularly concerning given the rapid advancements in text generation using large language models (LLMs) \cite{chen2023extensive, li2024chatgpt, tian2024opportunities}.  
Valuable applications of text generation in healthcare include generating detailed radiology report descriptions~\cite{tanida2023interactive} for more accurate diagnosis and distilling lengthy medical reports into concise summaries~\cite{van2023radadapt} for quicker decision-making. 
It also aids in creating personalized patient education materials~\cite{karabacak2023advent} and automating clinical trial protocols~\cite{subbiah2023next}, enhancing patient engagement and research efficiency.
 \textcolor{black}{As these applications become widely adopted \cite{weidinger2021ethical,miner2016smartphone, bickmore2018patient,sloan2024automated, pang2023survey,tian2024opportunities}, it is crucial to consider the unfairness problem and bias.}
For example, as the real example in Fig.~\ref{intro_case} shows, if a summarization model misses an important cardiomegaly observation from the doctor's findings for a male patient, this can lead to misdiagnosis or delayed treatment, potentially compromising patient care.
This raises an important question: \textbf{Does unfairness exist in AI-generated medical text, and if so, how can it be mitigated?}
Investigating this problem presents a greater challenge than a typical classification task, as generation is harder to evaluate, and maintaining fairness in the generation process is more difficult than simply outputting classification labels.

\begin{figure*}[!htb]
    \centering
    \includegraphics[width=1\linewidth]{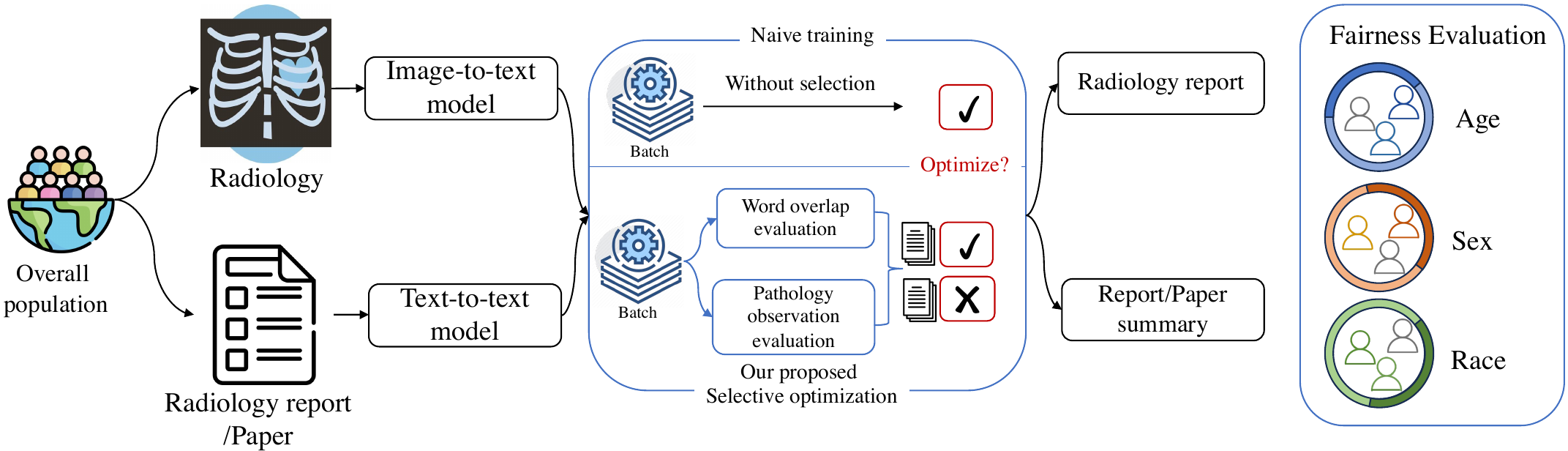}
    \caption{\textbf{Model Pipeline:} Our approach involves three tasks: radiology report generation, report summarization, and paper summarization, using the public benchmark MIMIC-CXR and PubMed datasets.
    We trained a deep-learning text generation model using our selective optimization mechanism, which considers two key aspects: word-level overlap and pathology observation.
    We evaluated pairwise fairness across different subgroups, including sex, race, and age, to determine if the model is equally fair for all individuals in each subgroup.}
    \label{intro}
\end{figure*}

In this study, we first evaluate the presence of unfairness issues in image-based computer-aided diagnosis, text-based radiology report, and medical paper summarization using publicly available datasets (Tab.~\ref{statistic}). 
Our evaluation spans six generation evaluation metrics and three different scale models, and considers both individual and intersectional groups across dimensions such as race, sex, and age.
We also propose a metric-aware unfairness indicator to evaluate the unfairness from different aspects.
Our experimental result shows that \textbf{the unfairness problem exists against certain groups}.
We also find that intersectional subgroups exhibit compounded biases in text generation, with patients who belong to two under-served subgroups receiving lower-quality diagnoses and experiencing larger discrepancies.
To address the issue of unfairness, we propose a novel selection optimization framework. 
Our first selection criterion relies on the intuitive cross-entropy loss function, where cases with higher loss in underrepresented groups are given more emphasis during the training process.
Apart from general quality considerations, we also want to consider domain-specific fairness enhancements.
There are metrics specifically designed for evaluating the accuracy of pathology concepts in medical text, and we prioritize training on cases that receive lower medical evaluation scores, thereby ensuring that the generated text precisely describes pathology terms.
We demonstrate that \textbf{our selective optimization can significantly mitigate unfairness across various metrics, datasets, and model scales for both individual and intersectional groups}, without compromising the model's overall performance. 
Our approach is not task-specific or model-specific, and can be applied to various areas of text generation, potentially effectively reducing bias issues.
An illustration of our model is presented in Fig.~\ref{intro}.

\begin{table}[!htp]
\centering
\begin{tabular}{l|l|c|c|c}
\toprule
\textbf{} & \textbf{Subgroup} & \textbf{Attribute} & \textbf{Total} & \textbf{Percentage} \\
\hline
\multirow{7}{*}[1em]{\rotatebox{90}{\parbox{3cm}{MIMIC Dataset}}}
 & \multirow{2}{*}{Age} & \textless 65 yrs & 46,336 & 54.67\% \\
 &  & $\geq$ 65 yrs & 55,902 & 45.32\% \\
\cline{2-5} 
 & \multirow{2}{*}{Sex} & Male & 50,273 & 49.17\% \\
 &  & Female & 51,965 & 50.82\% \\
\cline{2-5} 
 & \multirow{3}{*}{Race} & White & 64,783 & 63.36\% \\
 &  & Black & 19,568 & 19.13\% \\
\cline{2-5} 
 & \multirow{3}{*}{Split} & Train & 102,238 & 97.94\% \\
 &  & Val & 800 & 0.76\% \\
 &  & Test & 1,341 & 1.29\% \\
\midrule
\multirow{3}{*}[-1em]{\rotatebox{90}{Task}}
 & \multicolumn{2}{l|}{\textbf{}} & \textbf{Input} & \textbf{Output} \\
\cline{2-5} 
 & \multicolumn{2}{l|}{\multirow{1}{*}{Report Generation}} & \multirow{1}{*}{Images} & \multirow{1}{*}{Text} \\
 & \multicolumn{2}{l|}{\multirow{1}{*}{Report Summarization}} & \multirow{1}{*}{Text} & \multirow{1}{*}{Text} \\ \hline
  \midrule
\textbf{} & \textbf{Subgroup} & \textbf{Attribute} & \textbf{Total} & \textbf{Percentage} \\
\hline
\multirow{7}{*}[-3em]{\rotatebox{90}{\parbox{3cm}{PubMed Dataset}}}
 & \multirow{4}{*}{Age} & Adolescent & 6,749 & 42.29\% \\
 &  & Young Adult & 2,077 & 13.01\% \\
  &  & Middle Aged& 6,749 & 42.29\% \\
   &  &  Aged& 4,887 & 30.62\% \\
\cline{2-5} 
 & \multirow{3}{*}{Species} & Humans & 19,638 & 68.88\% \\
 &  & Animals & 6,906 & 24.22\% \\
\cline{2-5} 
 & \multirow{3}{*}{Split} & Train & 71,062 & 84.26\% \\
 &  & Val & 6,633 & 7.86\% \\
 &  & Test & 6,635 & 7.86\% \\
\midrule
\multirow{3}{*}[0em]{\rotatebox{90}{Task}}
 & \multicolumn{2}{l|}{\textbf{}} & \textbf{Input} & \textbf{Output} \\
\cline{2-5} 
 & \multicolumn{2}{l|}{\multirow{1}{*}{Paper Summarization}} & \multirow{1}{*}{Text} & \multirow{1}{*}{Text} \\
 \bottomrule
\end{tabular}
\caption{The characteristics of MIMIC-CXR and PubMed dataset for radiology report generation task, report summarization, and paper summarization task.}
\label{statistic}
\end{table}

\section{Evaluation Paradigm}

\subsection{Evaluation metrics}
\textcolor{black}{The evaluation of generated text quality is a complex and arduous topic, presenting greater challenges than the predominantly studied image classification task~\cite{zhang2019bertscore,celikyilmaz2020evaluation,fu2023gptscore}.} 
To reasonably compare performance across different groups, it is crucial to first determine how to evaluate the quality of the generation.
On the one hand, evaluation metrics based on $n$-gram overlap are the most commonly used and straightforward approach to assess text quality.
Previous studies indicate that metrics within this category, such as ROUGE~\cite{lin2004rouge}, exhibit a high correlation with human evaluation results~\cite{chen2024flexible}.
Other metrics like BERTScore~\cite{zhang2019bertscore}, ACU~\cite{Liu2022RevisitingTG}, and QuestEval~\cite{scialom2021questeval} also demonstrate comparable performance to ROUGE~\cite{chen2024rethinking}.
Therefore, in this paper, we choose ROUGE scores as one of the primary evaluation metrics.
Another paradigm of metrics is designed for radiology domains.
\textcolor{black}{Here, we use CheXpert scores~\cite{irvin2019chexpert}, a common method in the radiology field~\cite{wang2024cxpmrg,endo2021retrieval,boag2020baselines}, to evaluate generated text. Researchers define 14 chest pathologies as labels and assess the quality of the generated text by checking how well it detects and classifies these labels. The results are compared to ground truth labels. The labels include No Finding, Enlarged Cardiomediastinum, Cardiomegaly, Lung Opacity, Lung Lesion, Edema, Consolidation, Pneumonia, Atelectasis, Pneumothorax, Pleural Effusion, Pleural Other, Fracture, and Support Devices.}
Compared to ROUGE scores, CheXpert focuses more on the terms describing results of chest pathologies, rather than taking into account word overlap without differentiating the general and domain-specific terms.

Since previous work has not investigated the unfairness problem in text generation, in this study, we also introduce a Metric-Aware Fairness Difference (MFD) to quantitatively measure unfairness.
Inspired by the 'pairwise fairness difference' (PFD)~\cite{lin2023improving} in the classification domain, which subtracts the score of the lowest-performing group from the highest score within subgroups, our MFD adapts this concept for text generation. 
MFD is metric-aware, calculating a differential score between subgroups for each specific metric by subtracting the score of the lowest-performing group from the highest score within the subgroups. 
Compared with PFD, MFD can assess the degree of unfairness from various perspectives in the generated text.
\textcolor{black}{The formal calculation of MFD can be found in Section~\ref{metric}.}

\textcolor{black}{The justification for using MFD lies in its ability to capture unfairness across multiple dimensions relevant to text generation, which is inherently more complex than classification. 
In text generation, unfairness can manifest in subtle ways across various aspects of the generated output, such as stylistic inconsistencies or biases in content distribution. 
By providing a granular and metric-specific view of performance disparities, MFD facilitates a more comprehensive and targeted assessment of fairness in text generation systems.
A large MFD indicates significant disparities across subgroups, highlighting areas where fairness interventions are needed.}


\subsection{Compared models}
Our study investigates three tasks: radiology report generation, report summarization, and paper summarization. 
For the first task, we selected the specialized model R2Gen \cite{chen2020generating}, which is tailored for report generation. For the second task, we used the pre-trained language model BART-large \cite{lewis2020bart}. For the last task, we utilized both BART-base \cite{lewis2020bart} and the LLM  LLaMA2-13B \cite{touvron2023llama}. 
Our proposed paradigm is also applied to and compared with all the above models.
In this way, the sizes of the models we investigated range from 100M to 13B parameters, which provides a comprehensive investigation of fairness, and also allows us to thoroughly test our proposed method.

\begin{figure*}[!htb]
    \centering
    \includegraphics[width=1\linewidth]{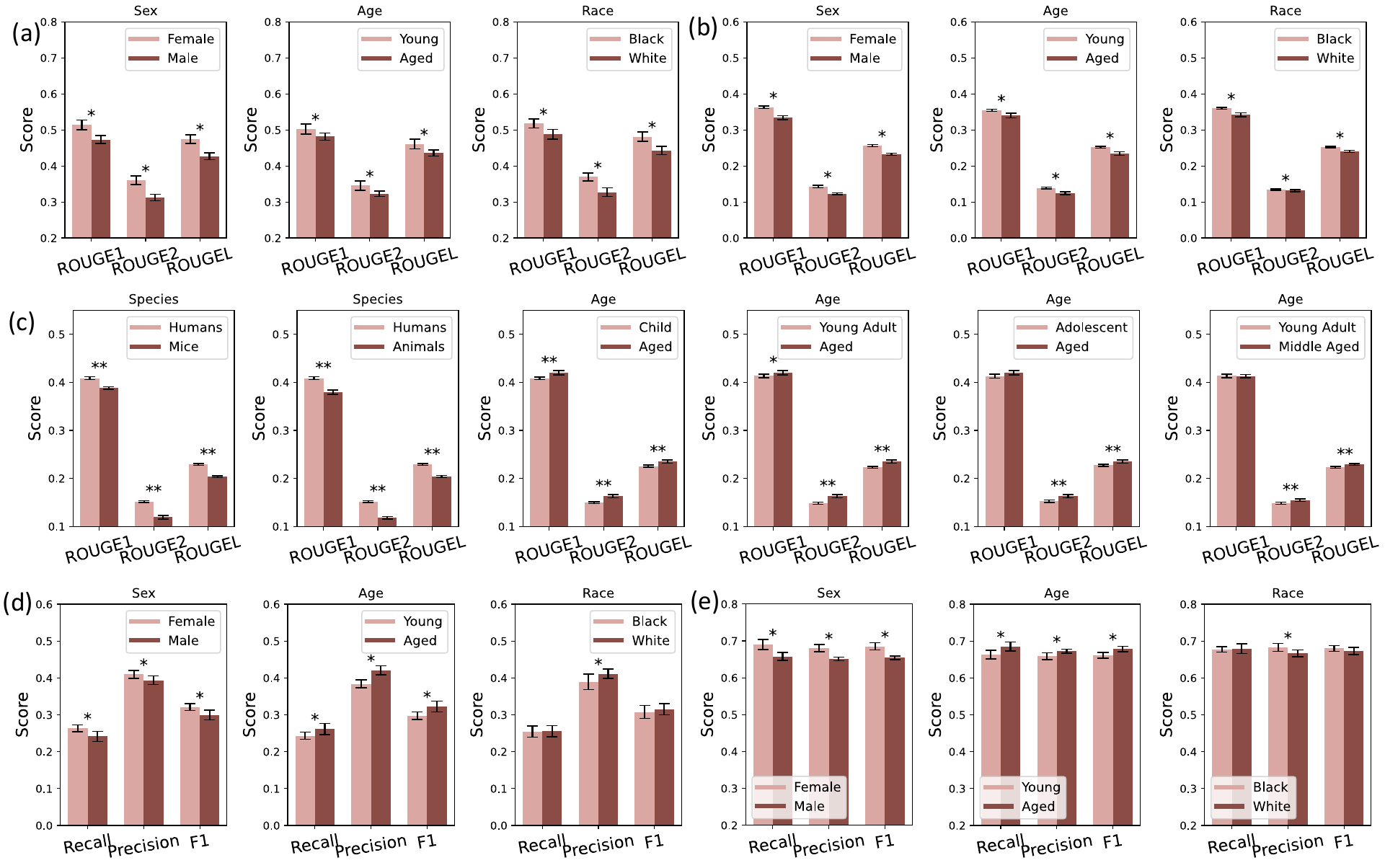}
    \caption{\textbf{Performance disparities were observed across various categories, including sex (female vs. male), race (black vs. white), age (young vs. aged), and species (humans vs. animals), for ROUGE comparison in: (a) radiology report generation, (b) summarization on the CXR dataset, and (c) summarization on the PubMed dataset; and for CheXpert comparison in: (d) radiology report generation and (e) report summarization.}
    Significant differences, denoted by * (p $<$ 0.05) and ** (p $<$ 0.01), were identified using the two-side Mann-Whitney U test.
    Error bars represent 95\% confidence intervals.}
    \label{red}
\end{figure*}

\subsection{Datasets}

We assessed the proposed and baseline models using the MIMIC-CXR \cite{johnson2019mimic} and PubMed~\cite{cohan2018discourse} datasets. 
\textbf{MIMIC-CXR}~\cite{johnson2019mimic} is a large public dataset of 377,110 chest X-rays associated with 227,827 free-text radiology report and summaries presenting to the Beth Israel Deaconess Medical Center Emergency Department between 2011 and 2016. 
With the release of its fourth version\cite{johnson2020mimic}, the dataset now includes corresponding patient information. 
The race and sex data are self-reported, and age is documented at the time of a patient's first admission.
We filtered out unpaired cases and accounted for instances where a single patient may have multiple X-rays and reports by randomly sampling one from each set. 
\textbf{PubMed}~\cite{cohan2018discourse} is a summarization dataset consisting of 133,215 full-text papers as documents and their abstracts as summaries. 
We collected MeSH labels using the PubMed API\footnote{\url{https://www.nlm.nih.gov/databases/download/mesh.html}}, resulting in 29,203 MeSH labels, and the evaluation is on the cases in test set with MeSH labels.
\textcolor{black}{Note that if a paper discusses both women and men, it is assigned to both categories.}

Supplementary Fig. S1 provides an overview of the data selection process.

\begin{figure*}[!htb]
    \centering
    \includegraphics[width=1\linewidth]{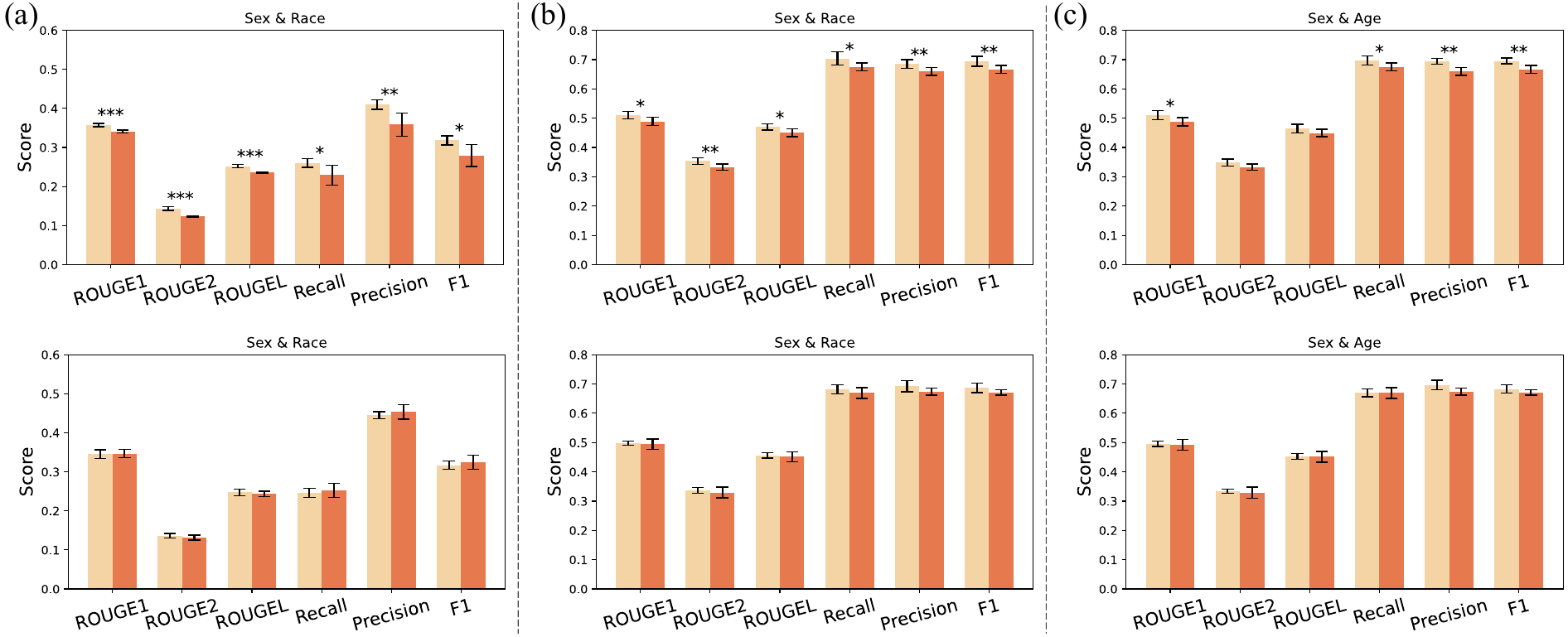}
    \caption{\textbf{Performance disparities were observed across intersectional groups in: (a) radiology report generation, and (b)(c) report summarization.}
    The top row represents baseline performance, while the bottom row illustrates the enhanced performance achieved by incorporating our proposed method.
    Significant differences, denoted by * (p $<$ 0.05), ** (p $<$ 0.01), and *** (p $<$ 0.001), were identified using the two-side Mann-Whitney U test. 
    Error bars represent 95\% confidence intervals.
    }
    \label{orange}
\end{figure*}

\section{Unfairness in existing models}

\subsection{Text generation bias in individual subpopulations on age, sex, and race}

As shown in Fig.~\ref{red}, we find that the text generation quality of baseline models for all datasets differs significantly in most of the considered subpopulations. 

Firstly, in Fig.~\ref{red}(a)(b)(c), we show the performance of different subgroups under ROUGE metrics across report generation, report summarization, and paper summarization, respectively.
For the first two tasks, it is observed that \textit{all} female, young, and black patients receive significantly higher quality summaries than their male, aged, and white counterparts.
This indicates that the generated text exhibits a higher word-level overlap with the ground truth reference. 
These findings are consistent across both tasks.
For the third task, the analysis includes more granular age comparisons as well as comparisons between humans and animals.
In the detailed age analysis, we observe that older individuals generally tend to receive better-performing summaries.
Additionally, significant differences are observed when comparing human data to animal data.

Next, in Fig.~\ref{red}(d)(e), we present the CheXpert scores of different groups on the first two tasks, which are based on the radiology domain. 
It is observed that, in 15 out of 18 settings, female, aged, and white individuals achieve higher scores.
This indicates that these groups receive higher quality generated results for medical conditions. 
Detailed performances are given in Supplementary Tables S1–7.

While the CheXpert score calculates the overlap of clinical observations between the generated summary and the reference summary, ROUGE scores consider all words.
Therefore, when we observe that females consistently achieve higher CheXpert scores and ROUGE scores than males, we also notice a distinct pattern: the generated text for young and black groups tends to score higher according to ROUGE metrics, whereas aged and white individuals score higher CheXpert scores. 
This discrepancy indicates that different metrics highlight different biases in model performance, making the alleviation of unfairness a challenging task that requires a multifaceted approach.

To offer a more intuitive understanding of the varying generation qualities, we present a representative example in Fig.~\ref{intro_case}, with additional cases detailed in Supplementary Fig.~S1-2. 
Despite similar inputs, males in these samples consistently received lower ROUGE and CheXpert scores, often resulting in missed or incorrect diagnoses. 
For instance, in the case depicted in Fig.~\ref{intro_case}, both female and male cases involve cardiomegaly and edema. 
Yet, the predicted impression for females accurately includes these observations, whereas the male prediction fails to acknowledge the cardiopulmonary condition.
It is crucial to emphasize that the observed biases could result in misdiagnoses or underdiagnoses for certain groups, potentially exacerbating existing health disparities. 
For example, if an AI system is biased against men or minorities, it might fail to accurately diagnose conditions that are more prevalent or present differently in these groups, such as cardiovascular diseases in men or skin conditions in people with darker skin tones. 
Addressing biases is crucial to ensure AI systems are equitable and provide fair, accurate assessments for all patients, regardless of age, sex, or race.

\subsection{Text generation bias in intersectional groups}

We next investigate intersectional groups, defined as patients belonging to two subpopulations, e.g., black female patients.
We highlight the subpopulations with the largest fairness disparities in intersectional groups like sex-race and sex-age in the first line of Fig.~\ref{orange}, with race-age comparisons shown in the supplementary Fig. S4.
Out of the 16/18 metric comparisons, significant differences are revealed, indicating that intersectional subgroups frequently experience notable biases in text generation.
To delve into the degree of bias, we compare the MFD between intersectional groups and subgroups in the Supplementary Tab. S9. 
This comparison indicates that patients belonging to two underserved subgroups are more likely to receive lower-quality diagnoses and experience greater discrepancies between groups.
For example, the disparity in CheXpert results between black males and white females is more pronounced than the disparities between black and white individuals or between females and males.
Detailed scores of Fig.~\ref{orange} are in the Supplementary Tab. S10.


\subsection{Why unfairness exists in radiology report generation tasks}

\textcolor{black}{While the previous section focused on identifying biases across different groups, it is equally important to understand why such unfairness exists. By uncovering the underlying causes, we can better address these disparities and develop more equitable models.}

Firstly, we find that the ROUGE score is related to the target length. The Pearson correlation between the ROUGE score and the reference length is -0.21 with a p-value of 3.90e-16, indicating a mild correlation where longer references tend to lead to lower ROUGE scores. 
This is intuitive because longer texts contain more information that needs to be generated, making the task more challenging.

Secondly, the CheXpert score is related to the original positive labels. 
If a group has more diseases classified as positive, its CheXpert score tends to be higher, with a correlation of 0.26 and a p-value of 9.79e-24. 
This indicates that the model tends to generate text mentioning pathologies rather than plain text without any disease mentions.
\textcolor{black}{Meanwhile, we observe that different demographic groups have varying probabilities of developing certain diseases. For example, Black patients have a 7\% higher likelihood of being diagnosed with pneumonia compared to White patients. Related works, such as \cite{mayr2010hospitals}, indicate that hospitals often provide lower-quality care to Black patients for pneumonia. This highlights the need to further explore disparities in disease prevalence and healthcare quality among different groups, which we leave for future work.}

Lastly, the number of training cases is also crucial. 
The correlation between different group case numbers and the ROUGE performance is 0.26 with a p-value of 0.02. 
This is also intuitive as a larger number of training cases provides more data for the model to learn from, leading to better performance.

It is important to note that a group's final performance is determined by multiple factors combined. 
Sometimes, one factor may outweigh the others.
For example, target text length is a decisive factor for ROUGE performance. 
However, in some cases, multiple factors work together to yield the final comparative outcomes. 
For instance, aged individuals have fewer training cases compared with young individuals (45\% compared to 54\%), more observation labels (5.13 compared to 5.01), and longer texts (62.89 words compared to 60.23 words).
The longer text factor and fewer training samples lead to worse ROUGE performance, but having more labels results in a higher CheXpert score.
\textcolor{black}{Due to the complexity of these distributions, achieving balance with a simple adjustment of one factor is challenging. 
To validate this, we performed an oversampling study, detailed in Supplementary Table S11, where we increased the number of male cases to match that of female cases.
However, the performance imbalance persisted, indicating that addressing such disparities requires more nuanced approaches.}

In summary, these factors contribute to an unstable and imbalanced performance across different groups, making the alleviation of these disparities a challenging task.

\section{Proposed framework mitigating the unfairness in medical text generation}

\subsection{Our proposed framework}

The pipeline of the proposed model is depicted in Fig.~\ref{intro}. 
The input images or text are passed into a neural network, which generates prediction results. 
The proposed method is versatile and not confined to specific deep-learning models.

Generally speaking, vanilla generation models are trained by considering all cases in a batch, processing them sequentially using cross-entropy loss to predict words one by one. In this context, we denote the cases within a batch as $B_i$, where $i$ represents the index number of the case.

In our approach, we introduce two novel paradigms for selecting cases for backpropagation, rather than using all cases.
Our first selection criterion is simple and intuitive: we prioritize cases that exhibit a larger cross-entropy loss. 
Formally, this can be expressed as:
\begin{align}
    B^*_{\text{selected}} = \text{Top } \gamma  \{ B_i \in B \mid \mathcal{L}_{\text{CE}}(B_i)\}.
\end{align}
Here, $\mathcal{L}_{\text{CE}}(B_i)$ denotes the cross entropy loss of case $B_i$, and $B^*_{\text{selected}}$ represents the subset of cases selected for backpropagation based on their higher loss values. 

\begin{figure}[!htb]
    \centering
    \includegraphics[width=1\linewidth]{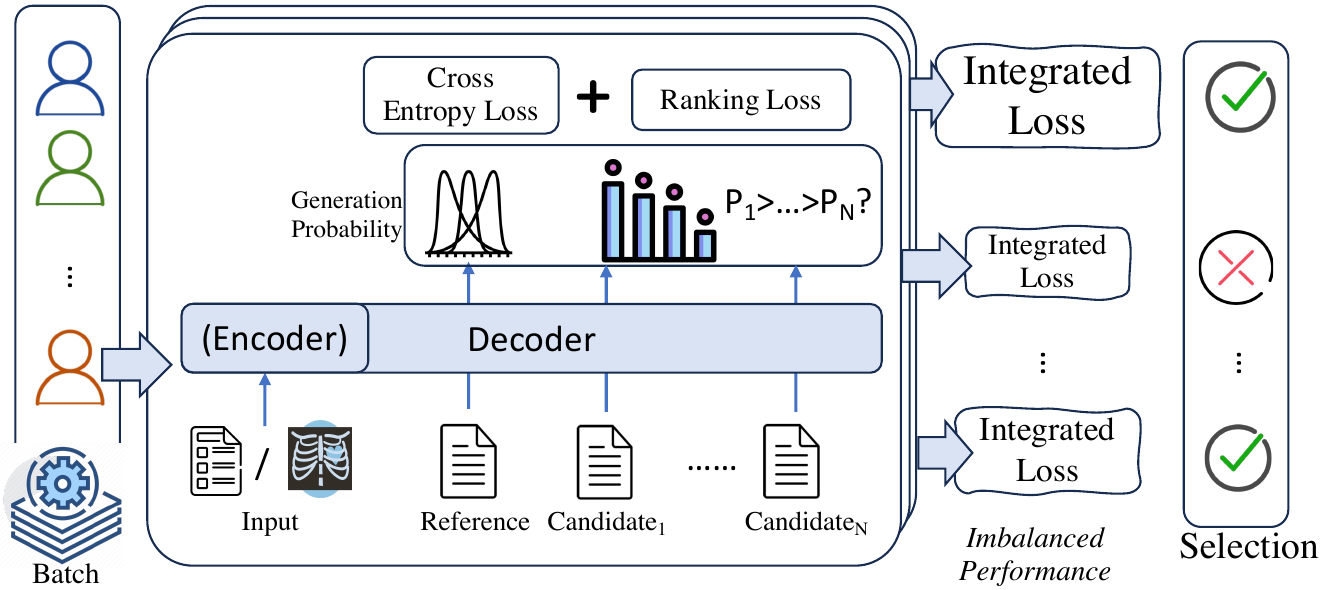}
    \caption{\textbf{The details of the selection algorithm.} The encoder-decoder or decoder-only model outputs the generation probability for the reference as well as a set of candidates. 
   The ranking loss evaluates if the model assigns a greater generation probability to candidates of higher quality compared to those of lower quality.
    Cases with a high sum of cross-entropy loss and ranking loss are selected for optimization. }
    \label{brio}
\end{figure}

The cross-entropy criterion emphasizes word-level accuracy. 
However, given our focus on medical applications, we also want to underscore the significance of accurately detecting pathology observations.
To address this, as depicted in Fig.~\ref{brio}, we modify the model to not only provide a prediction score for the ground truth reference but also for a set of reference candidates.
These candidates are generated by base models such as R2Gen, which are predefined and sorted according to their ROUGE and CheXpert scores, covering a range of quality levels.

The model then learns a ranking function that assigns higher prediction scores to candidates of higher quality. 
To achieve this, we employ a ranking loss that penalizes the model when it fails to rank high-quality candidates above lower-quality ones.
The ranking loss is defined as follows:
\begin{align}
\mathcal{L}_{\text{Ranking}} = \sum_{i=1}^{n-1} \max\left(0, \Delta_i - \text{score}(C_i) + \text{score}(C_{i+1})\right),
\end{align}
where $\mathcal{L}_{\text{Ranking}}$ is the ranking loss, $n$ is the number of candidates, $\Delta_i$ is the allowed margin between the scores of the $i$-th candidate $C_i$ and the $(i+1)$-th candidate $C_{i+1}$, and $\text{score}(C_i)$ is the model's prediction score for the $i$-th candidate.
The loss function encourages the model to learn that the score of the $i$-th candidate should be at least $\Delta_i$ higher than the score of the $(i+1)$-th candidate. 
If the model's predictions do not meet this criterion, the loss is non-zero and the model is penalized. 

In essence, for cases where the candidates are not ranked correctly, it indicates that the input case is more challenging and the model does not fully comprehend the input. Therefore, by integrating the ranking loss with the generation loss, we select the cases for training as follows:
\begin{align}
    B^*_{\text{selected}} = \text{Top } \gamma  \{ B_i \in B \mid \mathcal{L}_{\text{CE}}(B_i) + \mathcal{L}_{\text{Ranking}}(B_i) \}.
\end{align}
This combined approach ensures that the model is trained not only to generate accurate words but also maintain pathology accuracy, thereby improving the overall quality of the generated text.

\begin{figure*}[!htb]
    \centering
    \includegraphics[width=1\linewidth]{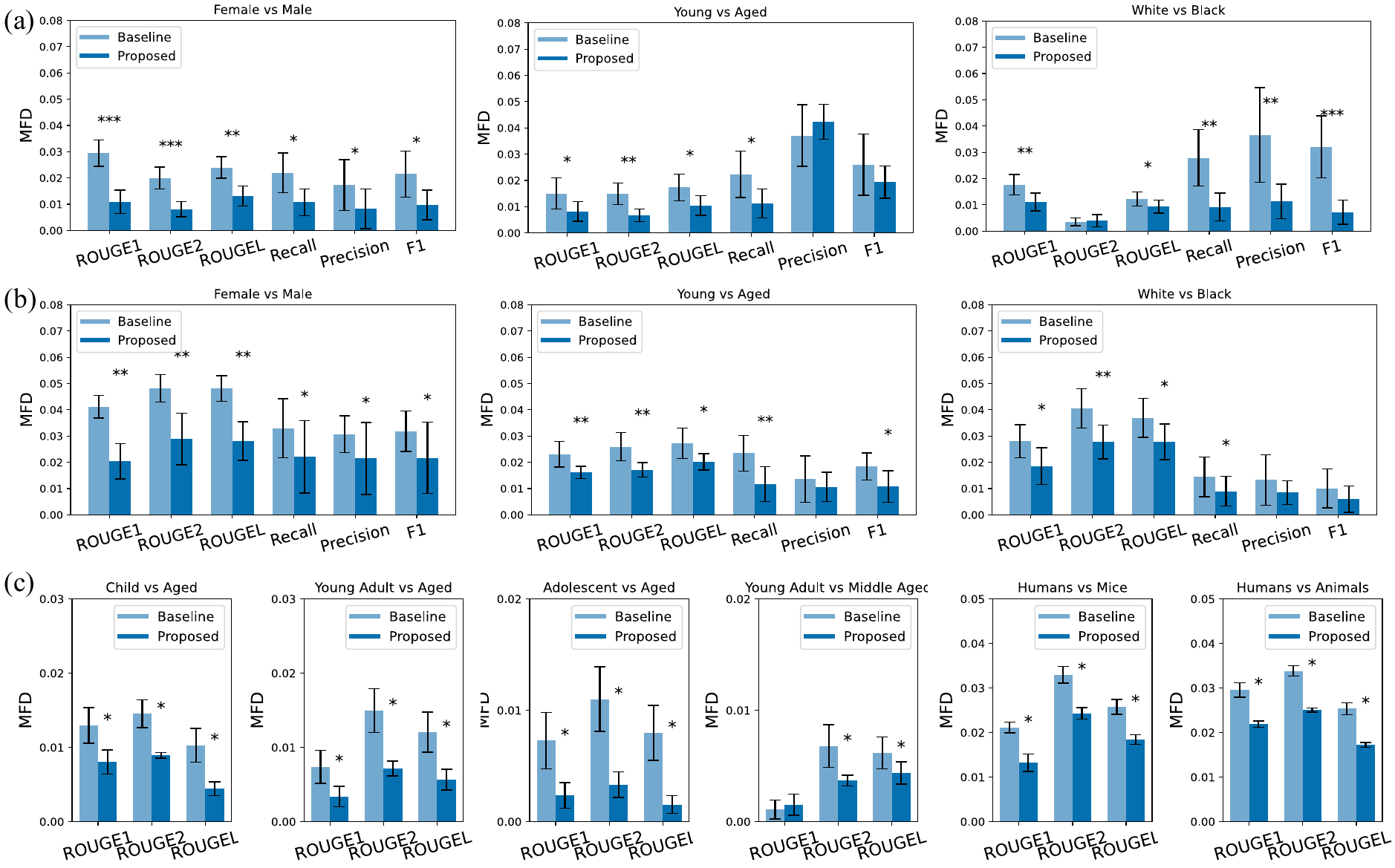}
    \caption{\textbf{Our proposed method significantly reduced the Metric-aware Fairness Difference (MFD) across various pairwise comparisons, such as female vs. male, as shown (a) radiology report generation, (b) report summarization, and (c) scholar paper summarization.}
    Significant differences, denoted by * (p $<$ 0.05), ** (p $<$ 0.01), and *** (p $<$ 0.001), were identified using the one-side Mann-Whitney U test. 
    }
    \label{black}
\end{figure*}

\subsection{Alleviating unfairness in subpopulations}

When equipping the baseline models with our proposed selective optimization, we find that our model is effective in reducing disparities across all datasets with respect to age, sex, and race.

In Fig.~\ref{black}(a)(b), we present the MFD score for radiology report generation and report summarization, respectively.
Across almost all metrics and tasks, the MFD scores of our model are significantly smaller than those of the baseline model, with the average MFD reduced by 35.27\%.
For the comparisons where the original discrepancy between groups is large, such as the female and male comparison, our model is particularly effective in alleviating bias. 
Moreover, although the discrepancies for different metrics vary—for example, black people have higher ROUGE scores but lower CheXpert scores compared with white people—our method is consistently useful for alleviating bias in these different metrics that measure different aspects.
This suggests that to improve fairness in other evaluation metrics, one can still adopt our framework.
In Fig.~\ref{black}(c), we demonstrate that our method is also effective in alleviating bias not only between human groups but also across different species.

\begin{figure*}[!htb]
    \centering
    \includegraphics[width=1\linewidth]{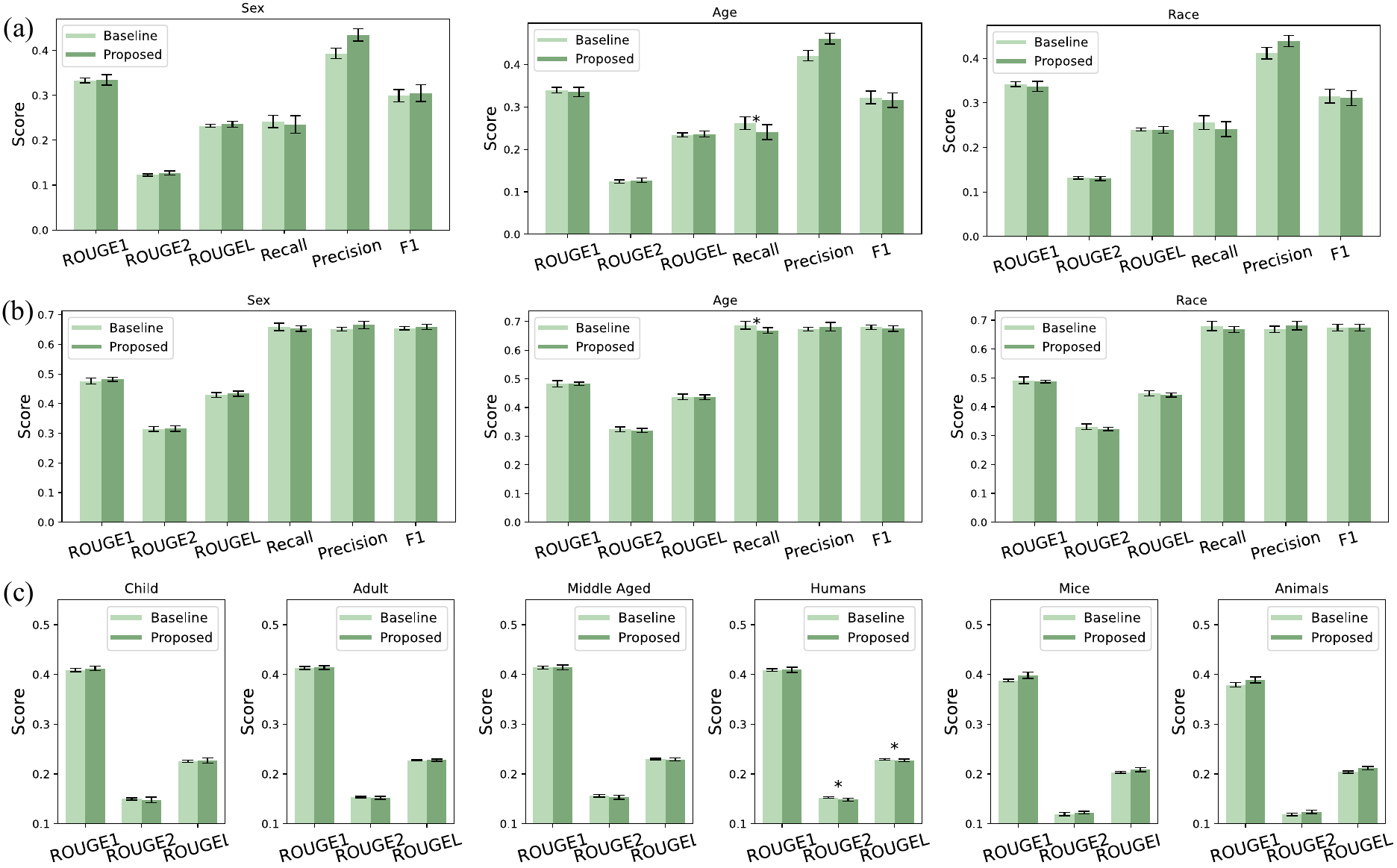}
    \caption{\textbf{
    Our proposed method maintains overall performance across different tasks: (a) radiology report generation, (b) report summarization, and (c) scholar paper summarization.} 
    Significant differences denoted by asterisks (*), were determined using the two-side Mann-Whitney U test. 
    Error bars represent 95\% confidence intervals.}
    \label{green}
\end{figure*}

While addressing bias, it is also crucial to maintain the quality of the original generated text. 
Otherwise, it would also bring harm if the performance of both groups significantly declines in the pursuit of fairness.
In Fig.~\ref{green}, we present a comparison between the original performance and the performance of our method. 
We also conduct the same significant test as before. 
As can be seen, for most of the time there is no significant performance drop between the two. 
Moreover, our method achieves better and more stable performance in some cases; for example, our method attains higher ROUGE scores, precision, and $F_1$ scores on the radiology report generation task across male, aged, and white groups, with a smaller 95\% error bar.
This demonstrates that achieving fairness can also enhance the overall performance and robustness of the model.
Details for the figures, overall performances, and comparison with baselines and other groups are given in Supplementary Tables S1–8 and Fig. S5.

\subsection{Alleviating unfairness in intersectional groups}

We also explore the effectiveness of our method for intersectional groups, as shown in Fig.~\ref{orange}.
We highlight the subpopulations with the largest fairness disparities in intersectional groups like sex-race and sex-age as shown in the three lower charts of the image, with race-age comparisons shown in the supplementary Fig. S4.
As observed, there is no longer a significant difference between comparisons in most metrics.
Our method even achieves superior performance, for example, on all three CheXpert metrics in the report generation task.
This suggests that our method, by maintaining balance, can indeed enhance the treatment received by intersectional groups, ensuring they benefit from improved outcomes.
Details of Fig.~\ref{orange} are in the Supplementary Tab. S10.

\subsection{Promoting fairness of LLMs generation}

\begin{figure*}[!htb]
    \centering
    \includegraphics[width=1\linewidth]{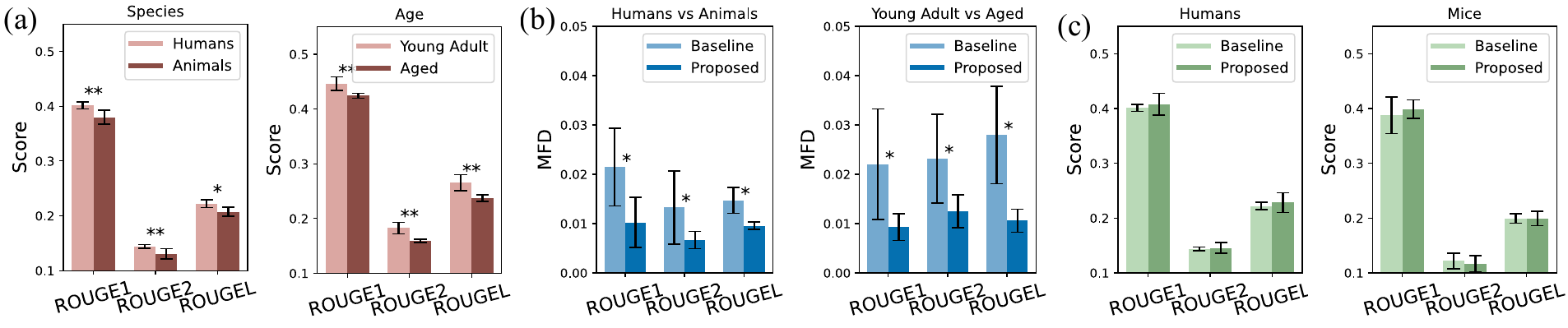}
    \caption{ 
   \textbf{(a) Significant performance differences were observed for the LLM Llama2-13B.
(b) Our method significantly reduced the MFD for Llama2.
(c) Our method does not adversely affect the overall performance of the Llama2.}
    }
    \label{mix}
\end{figure*}

In the previous experiments, we demonstrated the effectiveness of our proposed method across different model scales, from BERT-sized R2Gen to BART-large. 
In the era of LLMs, it is natural to wonder whether biases also exist within these models and whether selective optimization can be effectively applied to these larger models.
Herein, we choose Llama2-13B as our testbed.

In Fig.~\ref{mix}, we show (a) the original performance with significant differences, (b) the significant reduction in MFD using our method compared to the original MFD, and (c) the performance of three text generation metrics for the original and our method.
It is evident that significant performance disparities persist across various groups within LLMs, and our method continues to demonstrate effectiveness on models with billions of parameters, enhancing fairness while maintaining overall performance.
The full performance can be found in Supplementary Fig. S6 and Tab.S8.

The above results also show that unfairness in text generation is a common phenomenon even for LLMs trained on larger corpora. 
This underscores the importance of addressing biases to ensure fairness across different groups. 

\section{Methods}

\label{method}



\subsection{Evaluation metrics}
\label{metric}
We employ two types of evaluation metrics. 
The first is a set of traditional text generation metrics: ROUGE-1, ROUGE-2, and ROUGE-L, which respectively measure the matches of unigrams, bigrams, and the longest common subsequence. 
These metrics directly reflect the similarity between the generated text and the ground truth summary.

\begin{equation}
\text{ROUGE-1} = \frac{\sum_{S \in \text{References}} \sum_{\text{unigram} \in S} \text{Count}_\text{match}(\text{unigram})}{\sum_{S \in \text{References}} \sum_{\text{unigram} \in S} \text{Count}(\text{unigram})},
\end{equation}

\begin{equation}
\text{ROUGE-2} = \frac{\sum_{S \in \text{References}} \sum_{\text{bigram} \in S} \text{Count}_\text{match}(\text{bigram})}{\sum_{S \in \text{References}} \sum_{\text{bigram} \in S} \text{Count}(\text{bigram})},
\end{equation}

\begin{equation}
\text{ROUGE-L} = F_{\text{score}} = \frac{(1 + \beta^2) \cdot \text{R}_{\text{LCS}} \cdot \text{P}_{\text{LCS}}}{\text{R}_{\text{LCS}} + \beta^2 \cdot \text{P}_{\text{LCS}}},
\end{equation}
where
\begin{equation}
\text{R}_{\text{LCS}} = \frac{\text{LCS}(X, Y)}{\text{length of reference}},
\end{equation}
\begin{equation}
\text{P}_{\text{LCS}} = \frac{\text{LCS}(X, Y)}{\text{length of candidate}}.
\end{equation}
Additionally, we incorporate a specially designed metric, the CheXpert precision, recall, and F1 scores~\cite{irvin2019chexpert}, which automatically detect the presence of 14 observations in radiology reports, capturing the uncertainties inherent in radiograph interpretation. 
This metric has been shown to outperform at least two of the three radiologists in detecting four clinically relevant pathologies, demonstrating its capability in evaluating the accuracy of text.

From a fairness perspective, we introduce a Metric-aware Fairness Difference (MFD) metric, which computes the average absolute difference between two subgroups.
Formally, MFD is defined as follows:
\begin{align}
    \text{MFD} = \frac{1}{n} \sum_{i=1}^{n} | \text{Metric}_{\text{subgroup1}}(i) - \text{Metric}_{\text{subgroup2}}(i) |
\end{align}
Here, $n$ represents the number of instances, and $\text{Metric}_{\text{subgroup1}}(i)$ and $\text{Metric}_{\text{subgroup2}}(i)$ denote the metric values for the $i$-th instance in subgroup 1 and subgroup 2, respectively.
Being metric-aware, MFD is adept at capturing the actual disparities among various groups based on the inherent attributes of the metric itself, such as word-level accuracy or symptom detection accuracy.
This allows for a nuanced understanding of fairness in the context of the specific task at hand.

\textcolor{black}{The Mann-Whitney U test is a non-parametric statistical test used to determine whether there is a significant difference between two independent groups.
A smaller p-value indicates a significant difference, and better performance is determined by comparing the metric values of interest between the groups.}

\subsection{Experimental settings}
We conducted our experiments using Huggingface~\cite{wolf2020transformers} on NVIDIA A100 GPUs. 
For the R2Gen~\cite{chen2020generating} model, we employed a ResNet~\cite{he2016deep} pretrained on ImageNet~\cite{deng2009imagenet} as the visual extractor to extract patch features, with each feature having a dimension of 2,048. 
For the relational memory, we set the dimension to 512, the number of heads in multi-head attention to 8, and the number of memory slots to 3 by default. 
The model was trained using cross-entropy loss with the ADAM optimizer~\cite{kingma2014adam}. 
We set the learning rate to 5e-5 for the visual extractor and 1e-4 for other parameters. 
We decayed this rate by a factor of 0.8 per epoch for each dataset and set the beam size to 3 to balance generation effectiveness and efficiency.

For the BART model (facebook/bart-base and facebook/bart-large)~\cite{lewis2020bart}, we adhered to their hyperparameter settings as they yielded better performance.
We used the Adam optimizer with $\epsilon$ set to 1e-8 and $\beta$ set to (0.9, 0.999). 
The learning rate was set to 3e-5, with a warm-up of 500 steps. 
The batch size was set to 8, with 4 gradient accumulation steps.

For finetuning the LLM Llama2-13B, we used LoRA, which reduced the number of trainable parameters by learning pairs of rank-decomposition matrices while freezing the original weights. 
Specifically, we applied LoRA to the query projection layer and value projection layer to enhance the model's adaptability without significantly altering its structure. 
Additionally, we set the per-device training batch size to 16, utilized gradient accumulation with a step count of 1 to simulate larger batch sizes, and employed a cosine learning rate scheduler to optimize the learning rate adaptively throughout the training process.

\textcolor{black}{We also include an ablation study in Supplementary Table S12, where we remove the cross-entropy-based selection and the ranking-loss-based selection, respectively. 
The results demonstrate the effectiveness of both components in alleviating bias.}
All experiments in this paper were repeated at least five times following \cite{lin2023improving}. The average performance with 95\% confidence intervals is reported for each evaluation.

\noindent
\textbf{Data availability: }The MIMIC-CXR dataset used in this study is available in the PhysioNet database \url{https://www.physionet.org/content/mimic-cxr-jpg/}. 
The PubMed dataset is available at \url{https://huggingface.co/datasets/ccdv/pubmed-summarization}.
All data supporting the findings described in this manuscript are available in the article and in the Supplementary Information and from the corresponding author upon request. 
All source datasets are public datasets that can be accessed based on the links in this paper. 
Source data are provided with this paper.

\noindent
\textbf{Code availability: }The code is publicly available at \url{https://github.com/iriscxy/GenFair}.

\noindent
\textbf{Inclusion \& Ethics : }
In this study, we emphasize inclusivity and ethical responsibility in developing AI-driven text generation models for medical applications. 
Recognizing the risks of AI systems reflecting and amplifying societal biases, particularly in healthcare, we focused on addressing performance disparities across diverse demographic groups, including different races, sexes, and age groups. 
Our proposed algorithm specifically targets underperforming groups to reduce bias, ensuring that our models serve all populations equitably. 
We adhere to principles of fairness, transparency, and accountability, using only anonymized and publicly available data in compliance with privacy regulations. 
By making our code publicly accessible, we invite the broader research community to collaborate in enhancing the fairness and reliability of AI in healthcare.

\bibliographystyle{IEEEtran}
\bibliography{reg}

\end{document}